\documentclass[letterpaper, 10 pt, conference]{ieeeconf}  

\usepackage{cite}
\usepackage{amsmath,amsfonts}
\usepackage{algorithmic}
\usepackage{array}
\usepackage[caption=false,font=normalsize,labelfont=sf,textfont=sf]{subfig}
\usepackage{textcomp}
\usepackage{stfloats}
\usepackage{url}
\usepackage{verbatim}
\usepackage{graphicx}
\usepackage{multirow}
\usepackage[colorlinks,
            linkcolor=green,
            anchorcolor=green,
            citecolor=green]{hyperref}
\usepackage{balance}
\usepackage{siunitx}
\usepackage{amssymb}
\IEEEoverridecommandlockouts                              

\title{\LARGE \bf
Poly-MOT: A Polyhedral Framework For 3D Multi-Object Tracking 
}

\author{Xiaoyu Li$^\dagger$, Tao Xie$^\dagger$, Dedong Liu$^\dagger$, Jinghan Gao, Kun Dai, Zhiqiang Jiang, Lijun Zhao, Ke Wang 
\thanks{$\dagger$: These authors contributed equally to this work.}
\thanks{This work was in part by National Natural Science Foundation of China under Grant 62073101. (Corresponding authors: Lijun Zhao and Ke Wang.)}
\thanks{All authors are with State Key Laboratory of Robotics and System, Harbin Institute of Technology, Harbin 150006, China.}}

\begin{document}

\maketitle
\thispagestyle{empty}
\pagestyle{empty}

\begin{abstract}
3D Multi-object tracking (MOT) empowers mobile robots to accomplish well-informed motion planning and navigation tasks by providing motion trajectories of surrounding objects.
However, existing 3D MOT methods typically employ a single similarity metric and physical model to perform data association and state estimation for all objects. 
With large-scale modern datasets and real scenes, there are a variety of object categories that commonly exhibit distinctive geometric properties and motion patterns.
In this way, such distinctions would enable various object categories to behave differently under the same standard, resulting in erroneous matches between trajectories and detections, and jeopardizing the reliability of downstream tasks (navigation, etc.).
Towards this end, we propose Poly-MOT, an efficient 3D MOT method based on the Tracking-By-Detection framework that enables the tracker to choose the most appropriate tracking criteria for each object category.
Specifically, Poly-MOT leverages different motion models for various object categories to characterize distinct types of motion accurately. 
We also introduce the constraint of the rigid structure of objects into a specific motion model to accurately describe the highly nonlinear motion of the object.
Additionally, we introduce a two-stage data association strategy to ensure that objects can find the optimal similarity metric from three custom metrics for their categories and reduce missing matches. 
On the NuScenes dataset, our proposed method achieves state-of-the-art performance with 75.4\% AMOTA. The code is available at \href{https://github.com/lixiaoyu2000/Poly-MOT}{https://github.com/lixiaoyu2000/Poly-MOT.}

\end{abstract}

\begin{figure}[t]
      \centering
      \includegraphics[width=\linewidth]{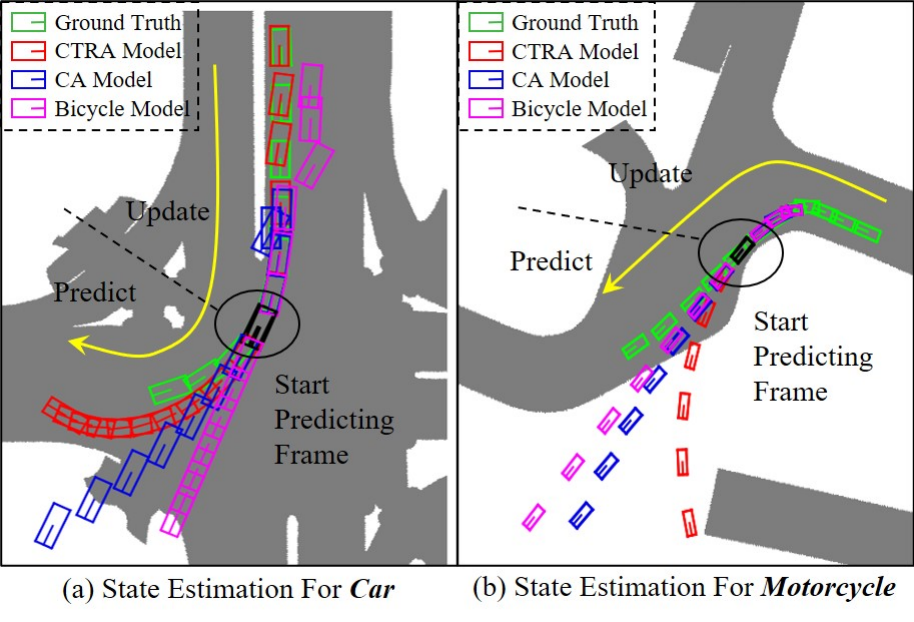}
      \vspace{-2.5em}
      \caption{\textbf{Trajectory state estimation of our proposed (\textit{CTRA} and \textit{Bicycle} Model) and existing (\textit{CA Model}) motion model on \textit{Car} and \textit{Motorcycle}.} For trackers equipped with different motion models, we truncate the tracking process from the same timestamp, which means that the trajectory can receive detection updates before this timestamp. In contrast, the trajectory can only use historical information to predict the future state after this timestamp. (a) \textit{CTRA Model} exhibits a significantly higher prediction accuracy for \textit{Car} than other models. This is particularly useful for recovering historical mismatch trajectories when objects are occluded, or detectors miss detections. (b) \textit{Bicycle Model} is more suitable for \textit{Motorcycle} due to different object categories exhibiting distinct motion patterns.}
      \vspace{-2em}
\label{fig:f0}
   \end{figure}


\section{INTRODUCTION}
Multi-Object Tracking (MOT) is a critical component of environment perception systems in autonomous robots.
It provides valuable information on the motion of tracked objects over time, enabling robots to predict the future motion patterns of surrounding objects effectively. Compared with 2D MOT~\cite{bytetrack, fairmot, deepsort}, 3D MOT~\cite{9341164} offers more explicit and convenient spatial information about objects, culminating in more reliable and accurate tracking. 
Typically, current 3D MOT techniques can be divided into ``Tracking-By-Detection" (TBD)~\cite{9562072, https://doi.org/10.48550/arxiv.2209.02540} and ``Joint Detection and Tracking" (JDT)~\cite{2022Minkowski, 9636311, 9578166}. 
Due to the data-driven nature of JDT, it is generally less precise and robust than TBD, and consequently, the majority of 3D MOT approaches adhere to the TBD architecture.

In the most previous works~\cite{2021SimpleTrack, 9341164, 9562072}, KITTI~\cite{geiger2012we} and MOT15~\cite{leal2015motchallenge} are employed to evaluate algorithm performance.  
Under these platforms, trackers are usually required to track only a single category of objects. 
Therefore, these works simply use a single linear motion model and similarity metric for state prediction and construct the cost matrix between trajectories and detections. 
However, with the advent of large-scale datasets such as NuScenes~\cite{caesar2020nuscenes} and changeable real scenes, a long-ignored yet fundamental fact must be carefully considered: 
\textit{there are multiple object categories in real scenes, and objects of different categories often exhibit various geometric features and motion patterns.} 
A single prediction and matching criterion is unsuitable for distinct object categories, which distorts the affinities between trajectories and detections, resulting in false matches and compromising the stability of subsequent tasks (navigation, prediction, etc.).

To the best of our knowledge, only a few recent works~\cite{9562072, https://doi.org/10.48550/arxiv.2209.02540} have optimized the MOT problem in multi-category settings. 
These methods prevent correlation between different categories by masking~\cite{https://doi.org/10.48550/arxiv.2209.02540} or removing~\cite{9562072} invalid costs in the cost matrix calculated under the same standard.
However, these methods can not tackle the issue of accurate tracking in multi-category settings fundamentally due to the \textit{inaccuracy of the cost matrix} induced by \textit{unreliable prediction} and \textit{irrationality metric}. 
On the one hand\footnote{In Fig. \ref{fig:f0}, \textit{CA} denotes \textit{Constant Acceleration}, \textit{CTRA} denotes \textit{Constant Turn Rate and Acceleration}}, as shown in Fig. \ref{fig:f0}, due to the distinct and nonlinear motion patterns of different object categories, utilizing the same linear motion model for state prediction will result in an unreliable estimation of motion. 
Moreover, due to variances in geometric features, different object categories are susceptible to various similarity metrics and correlation thresholds.
As presented in Table \ref{table:6}, we conduct a simple and intuitive experiment confirming that a single similarity metric cannot perform well in all object categories.
Thus, precise yet reliable motion prediction and affinity calculation for various object categories is a vital step toward deploying 3D MOT methods in real scenes.

\begin{table}
\vspace{0.5em}
\centering
\caption{Ablation studies using different similarity metrics for the distinct category (for \textit{Ped} and \textit{Bus}). AMOTA and IDS are reported best for different similarity metrics.}
\label{table:6}
\setlength{\tabcolsep}{4mm}{
\begin{tabular}{cccc}
\hline
\textbf{Category} &
\textbf{Similarity Metric} & \textbf{AMOTA$\uparrow$} & \textbf{IDS$\downarrow$}    \\\hline
\multirow{3}{*}{\textit{Ped}} & $gIoU_{3d}$       & 81.7  & 175  \\
                     & $gIoU_{bev}$               & 81.2  & 203  \\
                     & $d_{eucl}$                 & 81.0  & 220  \\\hline
\multirow{3}{*}{\textit{Bus}} & $gIoU_{3d}$       & 88.1  & 2     \\
                     & $gIoU_{bev}$               & 88.2  & 2   \\
                     & $d_{eucl}$                 & 87.5  & 1   \\\hline  
\vspace{-4em}
\end{tabular}}
\end{table}

To this end, we introduce Poly-MOT, a polyhedral framework for 3D MOT under multi object category scenes following the TBD framework.
Specifically, to ensure accurate motion prediction in such scenes, we introduce geometry constraints to the motion model and establish multiple motion models (\textit{CTRA} and \textit{Bicycle} model) based on the distinct features of each object category. 
For accurate object matching, we design three similarity metrics and then introduce categorical data association, in which the tracker selects the optimal similarity metric for different categories to achieve accurate affinity calculation.
We also employ a technique that combines Non-Maximum Suppression (NMS) and Score Filter (SF) to preprocess detections at each frame to eliminate the gap between detection task and tracking task. 
Finally, we additionally employ a combined count-based and confidence-based strategy so that Poly-MOT can handle the lifecycle of trajectories with various matching statuses. 

Poly-MOT is learning-free and not data-driven, using only detection results as input and achieving state-of-the-art performance and manageable real-time performance without substantial computational resources, as shown in Tables \ref{table:1} and \ref{table:2}. 
Thanks to the TBD framework, Poly-MOT achieves stable tracking performance with multiple 3D detectors (CenterPoint~\cite{9578166}, etc.). 
\textbf{With 75.4\% AMOTA, our technique achieves state-of-the-art performance on the NuScenes test set.} 
We anticipate that Poly-MOT can provide an effective 3D MOT baseline algorithm for the community. The primary contributions of this work are as follows: 

\begin{itemize}
\item We propose Poly-MOT, an efficient 3D MOT approach for multiple object category scenes based on the TBD framework.
\item We introduce geometry constraints to the motion model and establish multiple motion models (\textit{CTRA} and \textit{Bicycle} model) according to the distinct features of different object categories, enabling capture motion pattern differences between categories. 
\item We design three custom similarity metrics and a novel two-stage data association strategy to ensure that various objects can identify the optimal similarity metric for their categories, thus reducing missing matches.
\end{itemize}

\section{Related Work}

\textbf{3D Multi-Object Tracking}. 
Weng~\cite{9341164} pioneers the application of the TBD framework to the 3D MOT method, using Linear Kalman Filter and 3D IOU to build an advance and fast 3D MOT system. 
The TBD framework divides the tracker into four steps: 
(1) Receiving and preprocessing the 3D detection, 
(2) Predicting motion for active trajectories, 
(3) Correlating and matching trajectory with detection, 
(4) Managing the lifecycle of all state trajectories. 
SimpleTrack~\cite{2021SimpleTrack} applies simple techniques to analyze and improve each of these four parts, resulting in impressive tracking performance. 
EagerMOT~\cite{9562072} takes the lead in employing result-level fusion to integrate 2D and 3D detections, improving the robustness of tracker to false negatives from different sensor modalities. 
In addition to TBD, the JDT framework processes tracking and detection tasks in a single Neural Network(NN). 
Feature alignment between multiple modalities is an important yet difficult point of JDT. 

\textbf{Data Association in 3D MOT.}
Data association is the core of MOT, as it is accomplished by calculating a cost matrix between trajectories and detections with a similarity metric and then applying a matching algorithm to obtain the final associations. 
Geometry-based and appearance-based are two common types of similarity metrics.
The former leverages location and motion information to boost the performance under occlusion, and common metrics include 3D IOU~\cite{9341164}, 3D GIOU~\cite{https://doi.org/10.48550/arxiv.2209.02540, 2021SimpleTrack}. 
Appearance-based metrics, which utilize appearance information, can achieve more robust results in cases of large distance movement or low frame rate, as demonstrated in several studies~\cite{9636311,https://doi.org/10.48550/arxiv.2209.02540,xie2023deepmatcher}.
Multi-modal 3D MOT methods typically use multi-level correlation~\cite{https://doi.org/10.48550/arxiv.2209.02540, 9562072} (applying multiple metrics to match objects multiple times) to fuse different modalities and improve performance. 
Poly-MOT demonstrates the benefits of multi-level correlation in reducing FN matches in LiDAR-only methods. 
Hungarian algorithm~\cite{9341164, https://doi.org/10.48550/arxiv.2209.02540} and greedy algorithm~\cite{9578166} are commonly used to solve the cost matrix. 
A concern is that existing methods use a single similarity metric for all object categories, despite the differences in geometric and appearance features among them. 
In contrast, Poly-MOT enables the tracker to select the optimal metric for each category based on its characteristics. 

\textbf{Motion module in 3D MOT.} 
The motion module predicts the state of active trajectories, maintaining temporal consistency with detection.
Motion prediction techniques can be divided into learning-based and filter-based methods. 
The former usually uses NN to predict the inter-frame displacement. 
CenterPoint~\cite{9578166} uses a center-based detector to output 3D detections and predicts the displacement of objects between frames by adding a regression branch. 
Filter-based methods use real-world physical models for state transitions, exhibit better robustness and real-time performance, and are widely adopted by most methods. Kalman Filter is a widely used method. Most Filter-based methods typically use \textit{CA}~\cite{https://doi.org/10.48550/arxiv.2209.02540} or \textit{Constant Velocity (CV)}~\cite{9578166, 2021SimpleTrack, 9341164} model as the motion model. However, these models assume that the movements of objects on each coordinate axis are independent, ignoring nonlinear motion patterns constrained by geometry and differences in motion patterns across categories. 
Therefore, to
ensure accurate prediction in multi-category scenes,
we introduce geometry constraints and
establish multiple models based on distinct features
of each category.

   \begin{figure*}[t]
   \vspace{0.5em}
      \centering
      \includegraphics[width=1\textwidth]{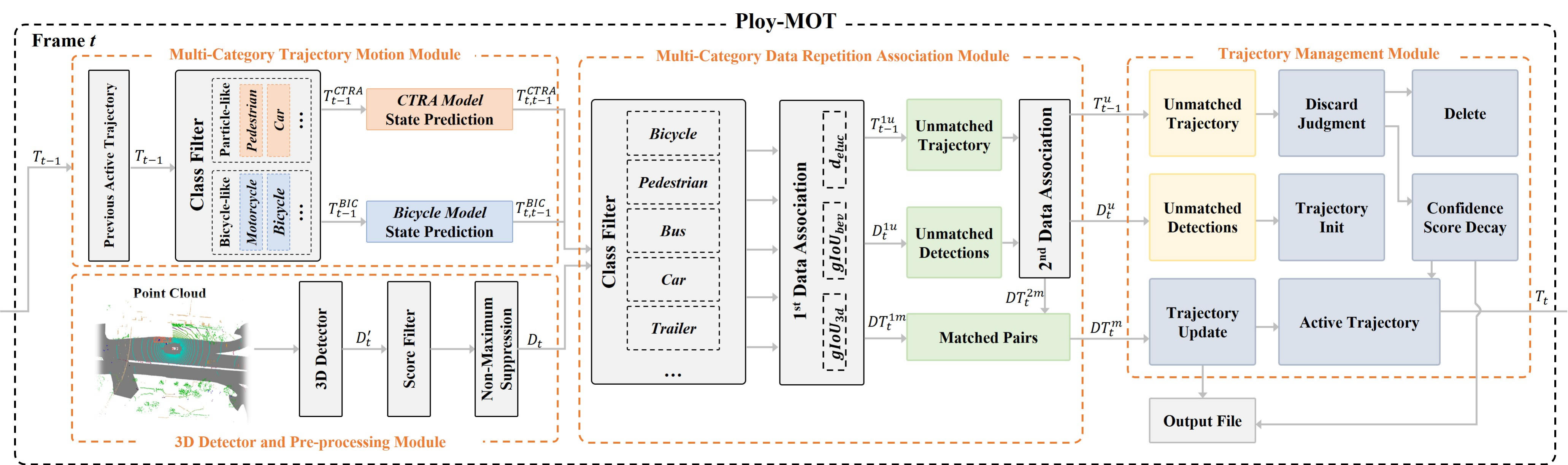}
      \caption{\textbf{The Pipeline Of Our Proposed Method At Frame $t$.} 
      (\uppercase\expandafter{\romannumeral1}) Previous active trajectory $T_{t-1}$ is divided into $T_{t-1}^{CTRA}$ and $T_{t-1}^{BIC}$ according to the different motion patterns. State predictions for $T_{t-1}^{CTRA}$ and $T_{t-1}^{BIC}$ are then made based on the distinct and nonlinear motion models using the EKF. 
      (\uppercase\expandafter{\romannumeral2}) Raw detections output by 3D detector are subjected to NMS and SF to reduce false positives to obtain $D_t$.
      (\uppercase\expandafter{\romannumeral3}) The prediction states $T_{t, t-1}^{CTRA}$, $T_{t, t-1}^{BIC}$, and $D_t$ are input to the Class Filter to classify the category. The first association is implemented within each category using the optimal similarity metric and a category-specific threshold. For unmatched trajectories $T_{t-1}^{1u}$ and unmatched detections $D_{t}^{1u}$, the second association is implemented with a distinct metric than before and a strict threshold. The final matched pairs $DT_{t}^{m}$ are used to update the corresponding trajectory. 
      (\uppercase\expandafter{\romannumeral4}) $D_{t}^{u}$ are initialized as new active trajectories. Part of $T_{t-1}^{u}$ are discarded based on the \textit{count-based} strategy, while others are added to the active trajectories again after the confidence score decays. Still active trajectories  will be output to the result file. Eventually, all active trajectories $T_{t}$ will be passed to the next frame $t+1$.}
      \vspace{-1.6em}
      \label{fig:f1}
   \end{figure*}

\section{Method}
Poly-MOT can be divided into four parts: the pre-processing module, multi-category trajectory motion module, multi-category data association module, and trajectory management module, as shown in Fig. \ref{fig:f1}.

\subsection{3D Detector and Pre-processing Module}
Existing 3D detectors~\cite{9578166,  largerkernel3d, xie2023poly} generate numerous low-confidence bounding boxes to ensure high recall, but applying these detections directly to update trajectories can result in severe ID switches (IDS). 
To tackle this issue, raw detections $D'_{t}$ must be preprocessed to reduce false-positive matches. 
We apply Non-Maximum Suppression (NMS) to $D'_{t}$ at each frame to remove bboxes with high similarity, improving precision without significant loss of recall. 
Nevertheless, each frame of the large-scale dataset (Waymo~\cite{sun2020scalability}, NuScenes~\cite{caesar2020nuscenes}, etc.) and real scenes usually contains a large number of objects while the number of $D'_{t}$ is significant. 
Directly applying NMS to $D'_{t}$ would lead to substantial computational overhead, as illustrated in Table~\ref{table:4}.
Before NMS, we apply a filtering process called Score Filter (SF) to remove detections $D'_{t}$ with confidence scores less than $\theta_{SF}$.
SF can efficiently remove apparent false-positive detections, improving the inference speed of the algorithm. 
After preprocessing, we obtain $D_{t}$, which includes the center of geometry position $(x, y, z)$, 3D size (width, length, height) $(w, l, h)$, heading angle $\theta$, and velocity $(v_x, v_y)$ on the ground plane. 
Note that whether velocity information is included or not depends on the dataset.


\subsection{Multi-Category Trajectory Motion Module}
Most previous methods~\cite{https://doi.org/10.48550/arxiv.2209.02540, 2021SimpleTrack} employ a uniform \textit{CA} or \textit{CV} model to predict the trajectories of all objects, whereas they fail to capture the highly nonlinear motion features of objects and ignore the differences in motion patterns across categories. 
To address this issue, we propose Multi-Category Trajectory Motion Module that utilizes different motion models (\textit{CTRA Model} and \textit{Bicycle Model}) for various object categories to characterize distinct types of motion accurately. 
In addition, we also introduce the constraint of the rigid structure of objects into a specific model to accurately describe the highly nonlinear motion of the object.
Notably, our motion models are formulated in the East(x)-
North(y)-Up(z) coordinate system, which follows the right-hand rule.

\begin{figure}[t]
   \vspace{0.5em}
      \centering
      \includegraphics[width=\linewidth]{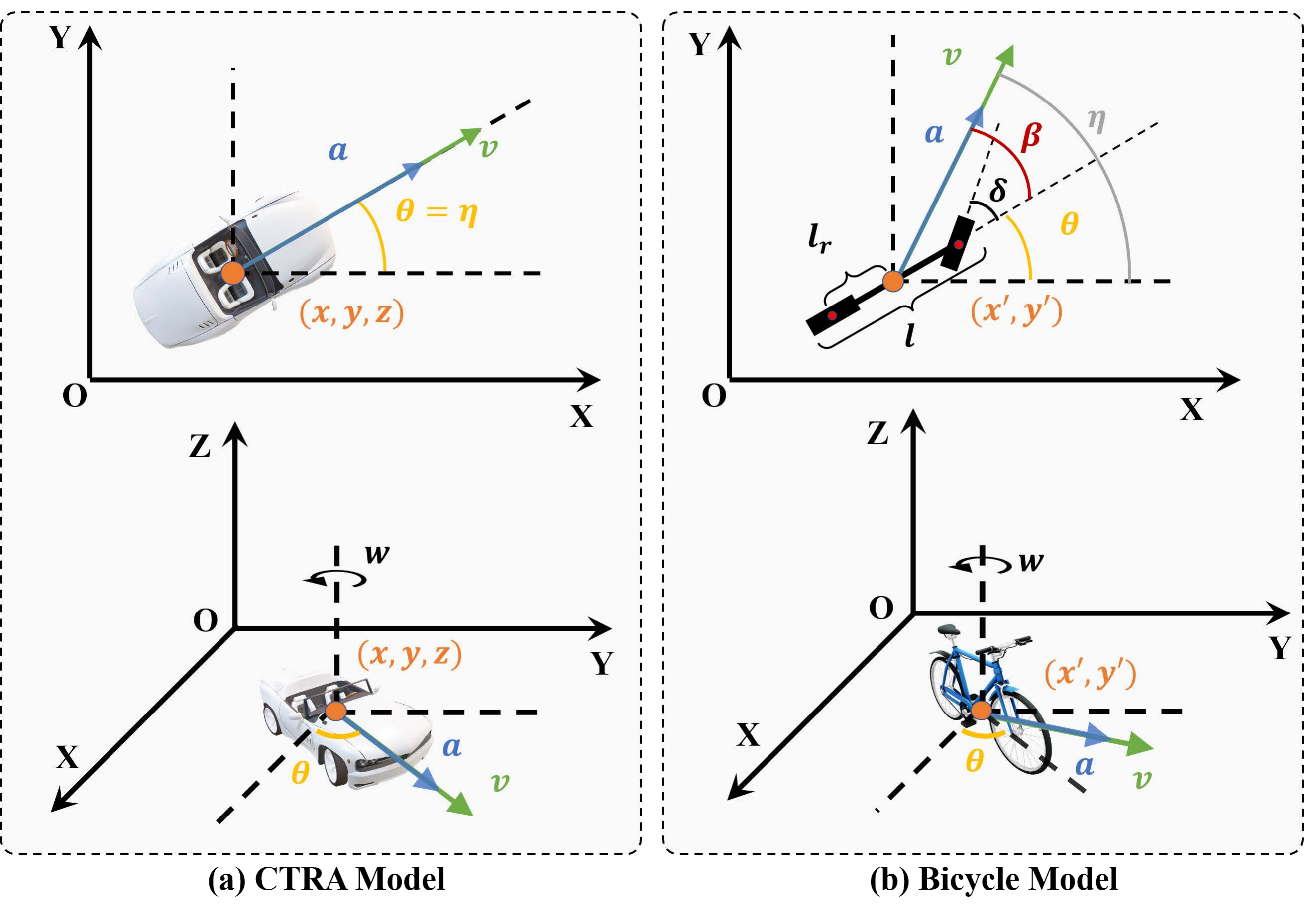}
      \vspace{-2em}
      \caption{\textbf{Representation of \textit{CTRA Model} and \textit{Bicycle Model} in 2D and 3D space.}}
      \vspace{-2em}
\label{fig:f2}
   \end{figure}

\textbf{CTRA Model}.  
For the \textit{CTRA model}, the turn rate $\omega$ and acceleration $a$ of the object are considered constant.
As shown in Fig. \ref{fig:f2} \textcolor{green}{(a)}, the heading angle and motion pattern of objects are tightly coupled in the \textit{CTRA model}, which means the directions of the heading angle $\theta$, velocity $v$, and acceleration $a$ of the object are on the same straight line.  
\textit{CTRA model} is suitable for \textit{car-like} objects and $pedestrian$. We formulate the state of an object trajectory as a 10-dimensional vector $T^{CTRA} = [x, y, z, v, a, \theta, \omega, w, l, h]$ in the \textit{CTRA model}, where $(x, y, z)$ represent the location of the geometric center of objects in the 3D space, $(w, l, h)$ represent the 3D size of objects. 

\textbf{Bicycle Model.}
For the \textit{Bicycle model}, it maintains the rigid structure of objects and enables the velocity direction and heading angle of objects to vary, rendering it suitable for objects that behave like bicycles, as illustrated in Fig. \ref{fig:f2} \textcolor{green}{(b)}. 
Meanwhile, we assume that the steering angle and velocity of the object remain constant. 
The state of the trajectory is also represented by a 10-dimensional vector $T^{BIC} = [x', y', z, v, a, \theta, \delta, w, l, h]$ 
, where $(x', y')$ represents the location of the gravity center of the object on the ground, $\delta$ represents the steering angle of the object, the remaining variables have the same meaning as the variables in $T^{CTRA}$.




\textbf{Model Establishment and State Prediction.}
Due to the nonlinear property of the motion models, we leverage the Extended Kalman Filter (EKF) to estimate the trajectory state. 
The prediction process can be described by: 
\begin{equation}
    T_{t,t-1} = f(T_{t-1}), \  P_{t, t-1} = F_{t}P_{t-1}F^{T}_{t} + Q, 
    \label{eq10}
\end{equation}
where $T_{t-1}$ denotes $T^{CTRA}$ or $T^{BIC}$, depending on the motion model of objects. 
$P_{t-1}$ is the covariance matrix at the previous moment $t-1$. 
$T_{t, t-1}$ is the predict state of $T_{t-1}$ at the current moment $t$. $Q$ is the process noise, which has an artificially set value. 
$f(\cdot)$ is the state transition function that is established from the motion model, reflecting the changes of all state variables of the trajectory between two consecutive frames. 
$F_{t}$ is the Jacobian matrix obtained through the partial derivative of $f(\cdot)$ with respect to $T_{t-1}$. 

During the state transition of all motion models, the variables $a, z, w, l, h$ are assumed to remain constant. 

The object location transition process as components of $f(\cdot)$ can be formulated as: 
\begin{equation}
    \hat{x}_{t, t-1} = \hat{x}_{t-1} + \int_{(t-1)\sigma}^{t\sigma}v(\tau)cos(\eta(\tau) )d\tau,  \label{eq12}
\end{equation}
\begin{equation}
    \hat{y}_{t, t-1} = \hat{y}_{t-1} + \int_{(t-1)\sigma}^{t\sigma}v(\tau)sin(\eta(\tau) )d\tau,  \label{eq13}
\end{equation}
where $\sigma$ is the interval between two adjacent frames of the LiDAR scan. Depending on the choice of motion model, the geometric center $(x, y)$ or gravity center $(x', y')$ of the object can be represented uniformly by $(\hat{x}, \hat{y})$. To better illustrate the state transition process of variables over time in each motion model, we introduce the time interval $\Delta t$, which is defined as follows:
\begin{equation}
    \Delta t=\tau - (t-1)\sigma.  \label{eq20}
\end{equation}
$\Delta t$ is the distance between the integral variable $\tau$ and the integral lower limit $(t-1)\sigma$ during the integration process. 
A tricky problem is that directly setting each variable in  \eqref{eq12} and \eqref{eq13} to be time-varying would result in non-integrable outcomes. 
A key insight is to leverage various motion models to simplify the complex nonlinear motion of objects to varying degrees, while accurately capturing the distinct motion patterns of different object categories.
The velocity transition function $v(\tau)$ is formulated as:
\begin{equation}
    v(\tau) =\begin{cases}
            v_{t-1}+a \Delta t  &\ \ if \ \ T = T^{CTRA} \\
            v_{t-1}             &\ \ if \ \ T = T^{BIC} \\
    \end{cases},
\label{eq14}
\end{equation}

Fig. \ref{fig:f2} illustrates the angle $\eta$ between the velocity of the object and the \textit{X-axis} of the coordinate system, and its state transition process is described by: 
\begin{equation}
    \eta(\tau) = \begin{cases}
                 \theta(\tau)    & \ \ if \ \ T = T^{CTRA}  \\ 
                 \theta(\tau)+\beta(\tau)   & \ \ if \ \ T = T^{BIC}  \\
    \end{cases},
\label{eq15}
\end{equation}
where $\beta$ represents the slip angle between the velocity and heading of the object, which can be calculated from assumed constant steering angle $\delta$ according to:
\begin{equation}
    \beta(\tau) = tan^{-1}(\frac{l_r}{\gamma l} tan(\delta(\tau))), \label{eq16}
\end{equation}
where $\gamma$ is the ratio of the wheelbase to object length $l$.
$l_r$ denotes the distance between the gravity center and the rear tire of the object, which is artificially set to 0.4-0.5 times the wheelbase. 
\eqref{eq16} is the embodiment of retaining the rigid structure of the object, and it also constitutes the major distinction between \textit{CTRA Model} and \textit{Bicycle Model}.
The reason for introducing $\beta$ is that the instantaneous center of the object in the \textit{Bicycle Model} is not on the body of the object.
In addition, incorporating $l$ in \eqref{eq16} signifies a deeper utilization of object observation and state information, enhancing motion accuracy.
However, a crucial observation that follows is that \textit{Bicycle Model} is susceptible to erroneous predictions caused by incorrect object structure information, thereby rendering it unsuitable for object categories where detectors tend to produce inaccurate detections.

$\theta(\tau)$ represents the heading angle transition function of an object, which is expressed uniformly in all models as: 
\begin{equation}
\theta(\tau) = \theta_{t-1} + \omega(\tau) \Delta t. \label{eq17}
\end{equation}
$\omega(\tau)$ in \eqref{eq17} describes the turn rate transition function, which is formulated as:
\begin{equation}
    \omega(\tau) = \begin{cases}
        \omega_{t-1} & if \ T = T^{CTRA} \\
        \frac{v(\tau)sin(\beta(\tau))}{l_r} & if \ T = T^{BIC}
    \end{cases},
    \label{eq19}
\end{equation}
which is actually constant in all motion models. \eqref{eq12}-\eqref{eq19} are the complete expression of state transition function $f(\cdot)$.



\subsection{Multi-Category Data Repetition Association Module} \label{sec:correlation}
In the data association process, a crucial but frequently disregarded fact exists: \textit{Different object categories are sensitive to various similarity metrics and association thresholds as a result of their unique geometric characteristics.}
However, most existing 3D MOT methods~\cite{2021SimpleTrack,9341164} leverage a single tracking standard for each category in multi-category scenarios, resulting in inferior tracking performance due to the lack of category-specific pertinence. 
To address these issues, we introduce Multi-Category Data Repetition Association Module that enables the tracker to choose the optimal similarity metric from a set of custom multiple metrics for each object category, thereby improving the accuracy and robustness of the MOT system. 
In addition, a two-stage association strategy based on different similarity metrics is applied to the module to reduce false negative matches.


\begin{table*}[]
\vspace{0.5em}
\caption[l]
{A comparison of existing algorithms applied to the NuScenes test set. The best performance is marked in \textcolor{red}{Red}, the second is marked in \textcolor{blue}{Blue}. (Bic, Motor, Ped, Tra, Tru) refers to (Bicycle, Motorcycle, Pedestrian, Trailer, Truck). }
\vspace{-2em}
\label{table:1}
\renewcommand{\arraystretch}{1.3}
\begin{center}
\setlength{\tabcolsep}{1.0mm}{
\begin{tabular}{cccccccccccccc}
\hline
\multirow{2}{*}{Method} & \multirow{2}{*}{Detector}  & \multirow{2}{*}{Input Data} & \multicolumn{8}{c}{AMOTA$\uparrow$}                                 & \multirow{2}{*}{IDS$\downarrow$} & \multirow{2}{*}{FP$\downarrow$} & \multirow{2}{*}{FN$\downarrow$} \\ \cline{4-11}
                        &                            &                             & Overall & Bic  & Bus  & Car  & Motor & Ped  & Tra  & Tru  &                      \\ \hline
CAMO-MOT~\cite{https://doi.org/10.48550/arxiv.2209.02540}                & BEVFuison~\cite{https://doi.org/10.48550/arxiv.2205.13542} \& FocalsConv~\cite{9880147}      & 2D + 3D                     & \textbf{\textcolor{blue}{75.3}}    & \textbf{\textcolor{red}{59.2}} & \textbf{\textcolor{blue}{77.7}} & \textbf{\textcolor{blue}{85.8}} & \textbf{\textcolor{blue}{78.2}}  & \textbf{\textcolor{red}{85.8}} & 72.3 & \textbf{\textcolor{red}{67.7}} & 324 & \textbf{\textcolor{red}{17269}} & \textbf{\textcolor{blue}{18192}}                 \\ 
CBMOT~\cite{9636032}                   & CenterPoint~\cite{9578166} \& CenterTrack~\cite{centertrack}   & 2D + 3D                     & 68.1    & 46.2 & 66.8 & 83.3 & 70.7  & \textbf{\textcolor{blue}{82.3}} & 69.6 & 57.5 & 709 & 21604 & 22828                 \\ 
EagerMOT~\cite{9562072}                & CenterPoint~\cite{9578166} \& Cascade R-CNN~\cite{8578742} & 2D + 3D                     & 67.7    & \textbf{\textcolor{blue}{58.3}} & 74.0 & 81.0 & 62.5  & 74.4 & 63.6 & 59.7 & 1156 & 17705 & 24925                 \\ 
Minkowski Tracker~\cite{2022Minkowski}             & Minkowski Tracker~\cite{2022Minkowski}               & 3D                     & 69.8    & 44.3 & 72.3 & 83.9 & 72.6  & 76.8 & \textbf{\textcolor{red}{75.3}} & 63.4 & 325 & 19340 & 21220                 \\ 
SimpleTrack~\cite{2021SimpleTrack}             & CenterPoint~\cite{9578166}                & 3D                          & 66.8    & 40.7 & 71.5 & 82.3 & 67.4  & 79.6 & 67.3 & 58.7 & 575 & 17514 & 23451                 \\ 
OGR3MOT~\cite{9691806}                 & CenterPoint~\cite{9578166}                & 3D                          & 65.6    & 38.0 & 71.1 & 81.6 & 64.0  & 78.7 & 67.1 & 59.0 & \textbf{\textcolor{red}{288}} & 17877 & 24013                 \\ 
CenterPoint~\cite{9578166}             & CenterPoint~\cite{9578166}                & 3D                          & 65.0    & 33.1 & 71.5 & 81.8 & 58.7  & 78.0 & 69.3 & 62.5 & 684 & \textbf{\textcolor{blue}{17355}} & 24557                 \\ \hline
Ours                    &LargeKernel3D~\cite{largerkernel3d}                  & 2D + 3D                      & \textbf{\textcolor{red}{75.4}}    & 58.2    & \textbf{\textcolor{red}{78.6}}  & \textbf{\textcolor{red}{86.5}}  & \textbf{\textcolor{red}{81.0}}  & 82.0   & \textbf{\textcolor{blue}{75.1}}  & \textbf{\textcolor{blue}{66.2}}  & \textbf{\textcolor{blue}{292}} & 19673 & \textbf{\textcolor{red}{17956}}                   \\ \hline
\vspace{-4em}
\end{tabular}}
\end{center}
\end{table*}

\textbf{First Association.}
After obtaining $T_{t, t-1}$ and $D_{t}$, affinity between $T_{t, t-1}$ and $D_{t}$ need to be calculated at each frame $t$. 
We first design three robust similarity metrics for distinct object categories to construct the first motion cost matrix $C^{1}_{t} \in R^{N_{cls} \times N_{det,t}\times N_{tra,t-1}}$ between $D_{t}$ and $T_{t, t-1}$. 
$N_{tra,t-1}$ and $N_{det,t}$ represent the number of $T_{t, t-1}$ and $D_{t}$, respectively. $N_{cls}$ is the number of categories in the dataset. 
We propose two similarity metrics \eqref{eq20}, \eqref{eq21}, \eqref{eq22} by the first time.
In addition, we introduce a rotation angle penalty factor in a specific metric to avoid false-positive associations in the opposite direction. 
These three similarity metrics, including 3D Generalized Intersection over Union ($gIoU_{3d}$), BEV Generalized Intersection over Union ($gIoU_{bev}$), and Euclidean Distance ($d_{eucl}$), are described as follows:

\begin{equation}
    gIoU_{3d}(B_{i}, B_{j}) = IoU_{3d}(B_{i}, B_{j}) + \frac{V(B_{i} \cup B_{j})}{V_{3dhull}(B_{i}, B_{j})} - 1, 
\end{equation}

\begin{equation}
    gIoU_{bev}(B_{i}, B_{j}) = IoU_{bev}(B_{i}, B_{j}) + \frac{A(B_{i} \cup B_{j})}{A_{bevhull}(B_{i}, B_{j})} - 1, \label{eq20}
\end{equation}

\begin{equation}
    d_{eucl}(B_{i}, B_{j}) = d(B_{i}, B_{j}) \ast (2-cos|\Delta \theta |), \label{eq21}
\end{equation}
\vspace{-1em}
\begin{equation}
    d(B_{i}, B_{j}) = \gamma_{geo}||B^{wlh}_{i}-B^{wlh}_{j}||_{2} + \gamma_{dis}||B^{xyz}_{i}-B^{xyz}_{j}||_{2}, \label{eq22}
\end{equation}
where $B$ is formulated as a high-dimensional vector representing the states of $T_{t, t-1}$ or $D_{t}$, which contain the 3D size and 3D center position. 
$IoU_{3d}$ and $IoU_{bev}$ are Intersection over Union in the 3D and bird's-eye view (BEV) representation space. 
$V(B_{i} \cup B_{j})$ and $A(B_{i} \cup B_{j})$ are the union volume and area of $B_{i}$ and $B_{j}$. 
$V_{3dhull}(B_{i}, B_{j})$ and $A_{bevhull}(B_{i}, B_{j})$ are the convex hulls computed by $B_{i}$ and $B_{j}$ in the 3D and BEV representation space. 
$B^{xyz}$ and $B^{wlh}$ are the vectors containing the 3D center position and 3D size of $B$. 
$\gamma_{geo}$ and $\gamma_{dis}$ are geometric and spatial distance ratios to the overall distance. 
$\Delta \theta \in \left [ 0,\pi  \right ]$ is the heading angle difference between $B_{i}$ and $B_{j}$. $|| \cdot ||_{2}$ is the 2-norm function.

For each category, we obtain the cost matrix $C^{1}_{t, cls} \in R^{N_{det,t} \times N_{tra,t-1}}$ by utilizing its optimal-performing similarity metric to compute the affinity of this category between $D^{cls}_{t}$ and $T^{cls}_{t, t-1}$\footnote{Costs between different categories are filled with invalid values\label{ft3}.}. 
After aggregating $C^{1}_{t, cls}$, we end up with $C^{1}_{t}$. 
Hungarian algorithm~\cite{kuhn1955hungarian} is employed to match $D_{t}$ and $T_{t, t-1}$ based on $C^{1}_{t}$. 
To account for the geometric size differences between objects of different categories, we employ different association thresholds $\theta _{fm} = \left ( \theta^{1}_{fm},\cdots,\theta^{N_{cls}}_{fm} \right )$ to constrain the matching process. 
After matching, we obtain three classes of matching instances, including matched pairs $DT^{1m}_{t} = \left \{ \left ( D^{i}_{t}, T^{j}_{t,t-1}\right ),\cdots  \right \}$, unmatched detections $D^{1u}_{t} \subseteq D_{t}$, and unmatched trajectories $T^{1u}_{t-1} \subseteq T_{t-1}$. $D^{1u}_{t}$ and $T^{1u}_{t-1}$ will be further associated in the second stage.

\textbf{Second Association.}
To reduce false-negative associations, we use $gIoU_{bev}$ for objects of all categories\footnote{If an object utilizes $gIoU_{bev}$ in the first association, then $gIoU_{3d}$ will be applied in the second stage, as the core of multi-stage association is to use different metrics to perform repeated associations.} to construct the cost matrix $C^{2}_{t} \in R^{N_{umdet,t}\times N_{umtra,t-1}}$ between $D^{1u}_{t}$ and $T^{1u}_{t-1}$ in the second stage\textsuperscript{\ref{ft3}}. $N_{umdet,t}$ and $N_{umtra,t-1}$ are the number of $D^{1u}_{t}$ and $T^{1u}_{t-1}$, respectively.  We use the Hungarian Algorithm with a strict threshold $\theta_{sm}$ based on the cost matrix $C^{2}_{t}$ to match $D^{1u}_{t}$ and $T^{1u}_{t-1}$. After aggregating the matching results of the two-stage association, we obtain the final matched pairs $DT^{m}_{t}$, unmatched detections $D^{u}_{t} \subseteq D_{t}$, and unmatched trajectories $T^{u}_{t-1} \subseteq T_{t-1}$.

\subsection{Trajectory Management Module} 
Following most 3D MOT methods~\cite{9562072, 9341164}, the trajectory management module is also responsible for four key functions, which include trajectory updating, trajectory initialization, trajectory death, and output file organization.
 
\textbf{Trajectory Update.} 
We utilize the detection in $DT^{m}_{t}$ and the standard update process of EKF to update the state of the corresponding trajectory and covariance matrix. 
It is important to note that in the state-measurement transition function $h(\cdot)$ of \textit{Bicycle model}, the geometric center of objects should be calculated based on the gravitational center.

\textbf{Trajectory Initialization.} 
We employ the \textit{count-based} approach to initialize $D^{u}_{t}$ as new tentative trajectories $T_{ten,t}$. 
If the $j$-th $T_{ten, t}$ is continuously hit in the next $hit_{min}$ frames, $T^{j}_{ten,t}$ will change to an activate trajectory and be merged into still active trajectories. 

\textbf{Trajectory Death.} 
We adopt the \textit{count-based} scheme to discard $T^{u}_{t-1}$.  Part of the trajectory in $T^{u}_{t-1}$ will be discarded if it has not been updated in the last \textit{max-age} frames. 
Trajectories that are not deleted are still considered active, but we penalize the confidence scores of these trajectories using $\alpha_{pun}$ and the exponential function $exp(\cdot)$.


\textbf{Result Output.} 
After obtaining all active trajectories $T_{t}$ at the current frame $t$, the updated trajectories (estimated motion state), newly initialized trajectories, and parts of the penalized trajectories are output to the result file. 
Note that, to reduce false-positive predictions, we only output $N_{pun}$ frames of the penalized trajectories' predicted state to the result file, and also apply NMS with $\theta_{nms}=0.08$ to all output trajectory states. 

\begin{table*}[t]
\vspace{0.5em}
\begin{center}
\caption[l]{
{A comparison of existing methods applied to the NuScenes val set. All main metrics reported in competitor papers are listed.}}
\vspace{-0.4em}
\label{table:2}
\renewcommand{\arraystretch}{1.3}
\setlength{\tabcolsep}{2.9mm}{
\begin{tabular}{cccccc}
\hline
\bf{Method} & \bf{Detector} & \bf{Input Data} & \bf{AMOTA$\uparrow$} & \bf{AMOTP$\downarrow$} & \bf{IDS$\downarrow$}  \\ \hline
CBMOT~\cite{9636032}   & CenterPoint~\cite{9578166} \& CenterTrack~\cite{centertrack} & 2D + 3D  & 72.0  & \textbf{\textcolor{red}{48.7}}  & 479   \\
EagerMOT~\cite{9562072}  & CenterPoint~\cite{9578166} \& Cascade R-CNN~\cite{8578742} &2D + 3D   & 71.2   & 56.9   & 899    \\
SimpleTrack~\cite{2021SimpleTrack}  & CenterPoint~\cite{9578166} &3D & 69.6  & 54.7  & 405  \\
CenterPoint~\cite{9578166}   & CenterPoint~\cite{9578166} & 3D & 66.5  & 56.7  & 562 \\
OGR3MOT~\cite{9691806}  & CenterPoint~\cite{9578166} &3D & 69.3  & 62.7  & \textbf{\textcolor{blue}{262}}  \\\hline
Ours      & CenterPoint~\cite{9578166} &3D  & \textbf{\textcolor{blue}{73.1}}   & \textbf{\textcolor{blue}{52.1}}  & 281   \\
Ours      & LargeKernel3D-L~\cite{largerkernel3d}  & 3D   & \textbf{\textcolor{red}{75.2}}    & 54.1   & \textbf{\textcolor{red}{252}} \\\hline
\vspace{-4.5em}
\end{tabular}}
\end{center}
\end{table*}

\section{EXPERIMENTS}

\subsection{Datasets}

\textbf{NuScenes.}
NuScenes~\cite{caesar2020nuscenes} contains 850 training sequences and 150 test sequences, each comprising approximately 40 frames showcasing diverse scenarios such as rainy days and nights. The keyframes are sampled at a frequency of 2Hz, and annotation information is provided for each keyframe. However, this keyframe frequency poses a challenge for precise motion model prediction, leading to significant inter-frame displacement. 
The official evaluator utilizes AMOTA as the primary evaluation metric~\cite{9341164}.

\subsection{Implementation Details}

\textbf{NuScenes.}
Our tracking method is implemented in Python under the Intel$^\circledR$ 9940X without any GPU. 
Hyperparameters are chosen based on the best AMOTA identified in the validation set. 
We utilize $\theta_{nms}=0.08$ for all categories and 3D detectors. 
$\theta_{SF}$ is detector-specific. $IoU_{bev}$ is used as the metric in NMS. 
During NMS process, objects of all categories are blended together. 
We employ \textit{Bicycle model} with $\gamma=0.8$ for \textit{(bicycle, motorcycle)} and \textit{CTRA model} for the remaining categories. 
The similarity metric for \textit{bus} and \textit{(bicycle, motorcycle, car, trailer, truck, pedestrian)} are $gIoU_{bev}$ and $gIoU_{3d}$, respectively. 
We apply $\theta_{fm}=(1.6,1.4,1.3,1.3,1.3,1.2,1.7)$ and $\textit{max-age}=(10,20,10,15,10,20,10)$ for \textit{bicycle, motorcycle, bus, car, trailer, truck, pedestrian} and $\theta_{sm}=1$ for all seven categories in the data association module.
For trajectory management, we set $hit_{min} = 0$, $\alpha_{pun}=0.05$, $N_{pun}=1$.  

\subsection{Experimental Results}

\subsubsection{Run-time discussion}
To solve the real-time challenge caused by extensive affinity calculations brought by a large number of objects, we first proposed the half-parallel\footnote{Since convex hull and rotation IoU calculations are still serial.} $gIoU$ operator under the Python implementation.
On the NuScenes, Poly-MOT can run at 3 FPS (Frame Per Second) on Intel 9940X, which has surpassed most advanced 3D MOT methods (SimpleTrack 0.51 FPS, Minkowski Tracker 1.7 FPS). 



\subsubsection{Comparative Evaluation}
We compare Poly-MOT to published and peer-reviewed state-of-the-art methods on the test and validation sets of the NuScenes dataset.

\textbf{NuScenes Test Set.}
Among all 3D MOT methods, Poly-MOT \textbf{ranks first} on the NuScenes tracking benchmark test set, i.e., 75.4\% AMOTA, exceeding most 3D MOT methods.
As shown in Table \ref{table:1}, Poly-MOT achieves an impressively low IDS 292 while maintaining the highest AMOTA (75.4\%) among all modal methods, which indicates that Poly-MOT is capable of achieving stable tracking without loss of recall.  
Without any image data as additional input, Poly-MOT still acquires state-of-the-art performance, surpassing the best-performing multi-modal tracker CAMO-MOT, which leverages a more superior integrated detector through~\cite{https://doi.org/10.48550/arxiv.2205.13542, 9880147}.
Additionally, Poly-MOT outperforms competing algorithms by a significant margin in the crucial category (\textit{Car}).
Compared to learning-based methods~\cite{2022Minkowski, 9578166, https://doi.org/10.48550/arxiv.2209.02540}, Poly-MOT incurs minimal computational overhead and delivers a more impressive performance, highlighting the promising potential of integrating filter-based 3D MOT methods into practical robotic systems. Notably, the IDS of Poly-MOT is slightly inferior to that of OGR3MOT~\cite{9691806}. However, the FN/FP in Table \ref{table:1} shows that Poly-MOT can offer the same robust continuous tracking capability without compromising recall.

\textbf{NuScenes Val Set.}
As presented in Table~\ref{table:2}, Poly-MOT outperforms other trackers in terms of both higher AMOTA and lower IDS when adopting the same detector (CenterPoint~\cite{9578166}).
Moreover, Poly-MOT yields an incredible tracking performance when assembled with a more strong LiDAR-only detector~\cite{largerkernel3d}, i.e., 75.2\% AMOTA, exceeding the best validation set accuracy reported by most methods.

\begin{table}[]
\vspace{0.5em}
\centering
\caption{The results of the ablation study of each module on the NuScenes val set. 
\textbf{OS} means original state. 
\textbf{Pre} means Pre-processing Module. \textbf{Mo} means Trajectory Motion Module. \textbf{Ass} means Data Association Module.}
\label{table:3}
\setlength{\tabcolsep}{1.8mm}{
\begin{tabular}{ccccc}
\hline
\textbf{Module} & \textbf{AMOTA$\uparrow$} & \textbf{IDS$\downarrow$} & \textbf{FN$\downarrow$} & \textbf{FP$\downarrow$}\\\hline
Os                  & 67.4   & 467    & 21442 & 14009\\
Os + Pre            & 71.4   & 374    & 18099 & 13299\\
Os + Pre + Mo       & 71.9   & 443    & 18086 & 13340\\
Os + Pre + Ass      & 72.0   & 410    & 15979 & 15932\\
Os + Pre + Mo + Ass & 73.1   & 281    & 17637 & 13437\\\hline
\vspace{-2.1em}
\end{tabular}}
\end{table}

\subsubsection{Ablation Studies}
In this part, we conduct extensive ablation experiments to evaluate the individual performance of proposed modules in Poly-MOT. 
We select CenterPoint~\cite{9578166} as the 3D detector and employ \textit{CA Model} with Linear Kalman Filter to predict the trajectory state from the Origin State (OS). 
We leverage $gIoU_{3d}$ and $\theta$ set to 0.14 as the similarity metric and association threshold, respectively. 
A series of experiments are then performed on the NuScenes validation set using various module combinations.

\begin{table}[]
\centering
\caption{The ablation study of whether or not to use Score Filter and Non-Maximum Suppression. Run-Time refers to the running time of Pre-processing Module.}
\label{table:4}
\setlength{\tabcolsep}{3.2mm}{
\begin{tabular}{cccc}
\hline
\textbf{Variable} & \textbf{AMOTA$\uparrow$} & \textbf{IDS$\downarrow$} &
\textbf{Run-Time (s) $\downarrow$}
\\\hline
NMS + SF & 73.1   & 281  & 0.055     \\
NMS      & 71.8   & 320  & 0.093     \\
SF       & 68.6   & 354  & 0.008     \\\hline
\vspace{-3.2em}
\end{tabular}}
\end{table}

\textbf{The effect of Pre-processing Module}. 
The significant gap between "Os" and "Os+Pre" in Table~\ref{table:3} showcases the impact of leveraging Pre-processing Module on the overall performance. 
We can observe that "Os+Pre" provides a +4\% AMOTA boost and a 93 IDS drop, resulting in a significant performance boost. 
The reason is that SF can filter out low-score bounding boxes while NMS can remove duplicate bounding boxes with high confidence, which makes the remaining bounding boxes have superior quality. 
In addition, using SF before NMS brings inference 40\% reduction in pre-processing inference time while boosting AMOTA by 1.3\% compared with only using NMS, as demonstrated in Table~\ref{table:4}. 

\textbf{The effect of Multi-Category Trajectory Motion Module.}
In Table~\ref{table:3}, we demonstrate the impact of the Multi-Category Trajectory Motion Module. 
"Os+Pre+Mo+Ass" achieves an AMOTA improvement of +1.1\% and an IDS decrease of 129 compared to "Os+Pre+Ass".
Benefiting from improved trajectory estimation, we can apply stricter thresholds to filter FP (-2495) in complex scenes (objects are dense and numerous, detectors exhibit poor performance, etc.) to achieve more stable tracking (-129 IDS) without incurring a significant loss in recall (+1658 FN). In addition, an intriguing observation is that while "Os+Pre+Mo" yields a +0.5\% AMOTA boost over "Os+Pre" alone, it also causes more ID switches (+69). 
The key insight is that the more accurate motion models change the bias distribution between predictions and ground truths for individual object categories, which makes a single metric and threshold unable to accurately capture inter-object affinities, thereby obtaining false matches and leading to IDS. 
Moreover, Table~\ref{table:5} reveals that using an inappropriate motion model for objects would decrease tracking performance, underscoring the importance of carefully deciding the motion model for each category.

\begin{table}[]
\vspace{0.5em}
\centering
\caption{Ablation studies using different motion models in Multi-Category (for \textit{Bic} and \textit{Moto}). AMOTA and IDS are reported best for different motion modules.}
\label{table:5}
\setlength{\tabcolsep}{2.2mm}{
\begin{tabular}{cccccc}
\hline
\textbf{Category} &\textbf{Motion Model} &\textbf{AMOTA$\uparrow$} & \textbf{IDS$\downarrow$} & \textbf{FP$\downarrow$} & \textbf{FN$\downarrow$}\\ \hline
\multirow{3}{*}{\textit{Bic}}    & \textit{Bicycle} & 57.1  & 0   & 227 & 765 \\ 
                            & \textit{CTRA}    & 55.4  & 0   & 256 & 747 \\  
                            & \textit{CA}      & 56.1  & 1   & 234 & 765 \\ \hline
\multirow{3}{*}{\textit{Moto}} & \textit{Bicycle} & 77.0  & 1   & 121 & 464 \\  
                            
                            & \textit{CTRA}    & 73.6  & 4   & 154 & 537 \\  & \textit{CA}      & 75.1  & 6   & 94 & 547 \\\hline
\vspace{-3.2em}
\end{tabular}}
\end{table}

\textbf{The effect of Multi-Category Data Repetition Association Module.}
As shown in Table~\ref{table:3}, "Os+Pre+Mo+Ass" achieves a +1.2\% AMOTA improvement and a -162 IDS reduction compared to "Os+Pre+Mo". This shows our proposed two-stage categorical association strategy can better capture the affinity between tracklet and detection of each category, enabling a more accurate matching relationship, improved tracking results and reduced FN matches. 

\begin{figure*}[t]
    \vspace{0.5em}
  \centering
  \includegraphics[width=1\textwidth]{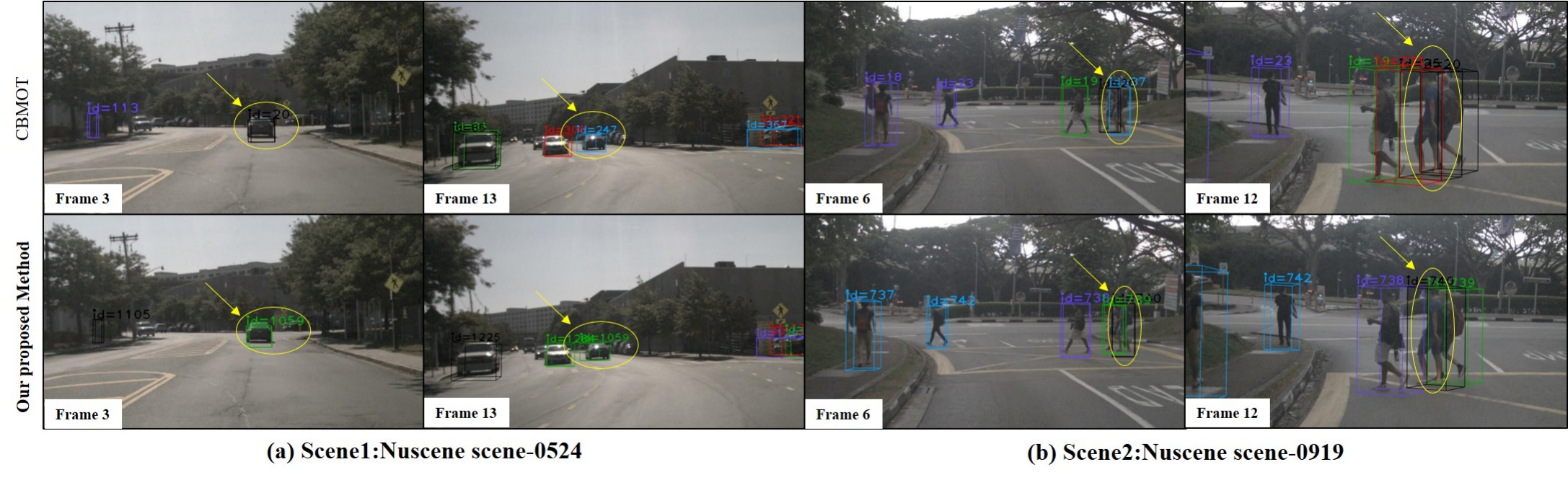}
  \vspace{-2em}
  \caption{\textbf{Visualization of comparison results between \textbf{Poly-MOT} and CBMOT~\cite{9636032}}. All methods use CenterPoint as a 3D detector. CBMOT simultaneously uses CenterTrack~\cite{centertrack} as a 2D detector for multi-modal fusion.}
  \vspace{-1.5em}
  \label{fig:f3}
\end{figure*}

\subsection{Visualization}
We qualitatively compare our Poly-MOT (LiDAR-only version) and advanced multi-modal 3D MOT method CBMOT on the NuScenes val set. 
As shown in Fig. \ref{fig:f3} \textcolor{green}{(a)}, when the object moves intensely and quickly, CBMOT has ID switches (ID changes from \textit{20} to \textit{247}), while the Poly-MOT can still achieve stable tracking.
As shown in Fig. \ref{fig:f3} \textcolor{green}{(b)}, when objects are dense and have irregular movement, CBMOT not only has ID switches (ID changes from \textit{37} to \textit{25}) but also fails to effectively suppress false-positive detection (ID: \textit{231} at Frame 12), while Poly-MOT still maintains stable tracking. 
The above comparison results show that Poly-MOT can alleviate the problem that LiDAR-only trackers cannot accurately track objects with large inter-frame displacements. 
In addition, Poly-MOT can also achieve stable tracking when the object suffers from occlusion.

\section{CONCLUSIONS}
In this work, we introduce Poly-MOT, a polyhedral framework for 3D MOT under multi object category scenarios following the TBD framework. 
Poly-MOT achieves accurate matches between tracklets and detections in multi-category scenarios by ensuring prediction reliability and metric rationality, including: 
(1) Two distinct and nonlinear motion models (\textit{CTRA} and \textit{Bicycle} Model) are established to represent the motion patterns of different object categories;
(2) Three similarity metrics ($gIoU_{3d}$, $gIoU_{bev}$, $d_{eucl}$) are designed to calculate the affinity of different object categories.
Besides, a two-stage association strategy and confidence-based pre-processing module are applied to the tracker to reduce FN matches and eliminate the gap between detection and tracking.
Without requiring additional training and GPU, Poly-MOT achieves state-of-the-art tracking performance with 75.4\% AMOTA on the NuScenes dataset while achieving an impressive inference speed. 
Our method can be easily combined with multiple detectors, and we envision it serving as a general baseline for future 3D MOT methods.


\addtolength{\textheight}{-12cm}   




\end{document}